\documentclass{article}
\PassOptionsToPackage{numbers, compress, sort}{natbib}

 \usepackage[dblblindworkshop, final]{neurips_2025}
\workshoptitle{Foundation Models for the Brain and Body}

\usepackage[utf8]{inputenc} 
\usepackage[T1]{fontenc}    
\usepackage{hyperref}       
\usepackage{url}            
\usepackage{booktabs}       
\usepackage{amsfonts}       
\usepackage{xcolor}         

\usepackage{subcaption}     
\usepackage{tabularx}
\usepackage{booktabs}       
\usepackage{multirow}
\usepackage{array, makecell}
\usepackage{rotating}
\usepackage{tabu}
\usepackage{amsmath,amssymb}
\usepackage{bbm}
\usepackage{nicefrac}       
\usepackage{microtype}      
\usepackage{pifont}
\usepackage{bbding}
\usepackage{graphicx,verbatim}

\title{Are foundation models useful feature extractors for electroencephalography analysis?}

\author{%
  Özgün Turgut\textsuperscript{\ensuremath{1}}
  \quad
  Felix S. Bott\textsuperscript{\ensuremath{2}}
  \quad
  Markus Ploner\textsuperscript{\ensuremath{2,3}}
  \quad
  Daniel Rueckert\textsuperscript{\ensuremath{1,4,5}}\vspace{.2cm} \\
  \textsuperscript{\ensuremath{1}}{Chair for AI in Healthcare and Medicine, Technical University of Munich, Germany}\\ 
  \textsuperscript{\ensuremath{2}}{Department of Neurology, Technical University of Munich, Germany}\\
  \textsuperscript{\ensuremath{3}}{Center for Interdisciplinary Pain Medicine, Technical University of Munich, Germany}\\
  \textsuperscript{\ensuremath{4}}{Munich Center for Machine Learning, Munich, Germany}\\
  \textsuperscript{\ensuremath{5}}{Department of Computing, Imperial College London, United Kingdom}\\
  \{\texttt{oezguen.turgut, daniel.rueckert}\}\texttt{@tum.de}
}

\begin{document}

\maketitle

\begin{abstract}
The success of foundation models in natural language processing and computer vision has motivated similar approaches in time series analysis. 
While foundational time series models have proven beneficial on a variety of tasks, their effectiveness in medical applications with limited data remains underexplored. 
In this work, we investigate this question in the context of electroencephalography (EEG) by evaluating general-purpose time series models on age prediction, seizure detection, and classification of clinically relevant EEG events.
We compare their diagnostic performance against specialised EEG models and assess the quality of the extracted features. 
The results show that general-purpose models are competitive and capture features useful to localising demographic and disease-related biomarkers.
These findings indicate that foundational time series models can reduce the reliance on large task-specific datasets and models, making them valuable in clinical practice.
\end{abstract}

\section{Introduction}
Recent breakthroughs in natural language processing and computer vision have shown the effectiveness of foundation models on a wide range of tasks. 
Inspired by this success, a growing number of works has focused on developing similar models for time series analysis \cite{Das2024,Gao2024,Goswami2024,Jiang2024,Turgut2025b,Woo2024,Yang2024,Liu2025,Shi2025}. 
While most of the foundational time series models are designed for only a single task such as forecasting \cite{Das2024,Woo2024,Liu2025,Shi2025}, 
recent works \cite{Gao2024,Goswami2024,Turgut2025b} have introduced general-purpose models that are effective on diverse tasks, including regression, classification, and forecasting. 
This raises the questions of (1) whether and (2) how \emph{general-purpose models} can be \emph{translated into a medical context} to benefit clinical applications with limited data availability. 

One relevant clinical application is electroencephalography (EEG), a widely accessible and cost-effective tool for measuring electrical brain activity across frequencies ranging from $0.5$ to $100\,$Hz, typically grouped into unified bands: delta ($\delta$: $0.5$–$4\,$Hz), theta ($\theta$: $4$–$8\,$Hz), alpha ($\alpha$: $8$–$13\,$Hz), beta ($\beta$: $13$–$30\,$Hz), and gamma ($\gamma$: $30$–$100\,$Hz) \cite{Newson2019}. 
Despite its accessibility, large-scale EEG datasets ($>$$10\,$k samples) remain scarce, limiting the ability to learn generalisable EEG features.
As a result, existing EEG models are typically task-specific, with architectures and training schemes tailored to applications such as age prediction \cite{Babayan2019,Engemann2022}, seizure detection \cite{Andrzejak2001}, or event type classification \cite{Harati2015}.
This task-specific nature leads to poor model generalisability and requires substantial effort to design and train new models for each use case.
In contrast, clinicians might tackle diverse EEG use cases more effectively using general-purpose models, given their general time series knowledge obtained through pre-training on large, heterogeneous datasets ($>$$100\,$k samples).

\begin{figure}[t]
    \centering
    \includegraphics[width=1.0\linewidth]{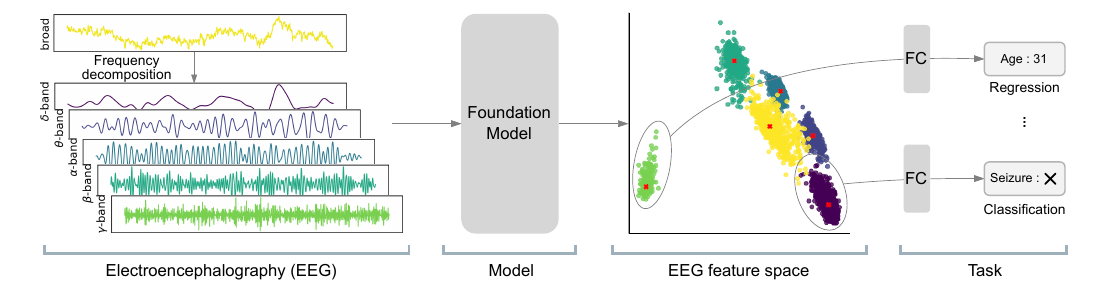}
    \caption{
    \textbf{Overview}. 
    We study the ability of general-purpose time series models to extract meaningful electroencephalography (EEG) features for tasks such as age prediction or seizure detection.
    }
    \label{fig:abstract}
\end{figure}
To this end, we systematically investigate the applicability of general-purpose models to EEG analysis (see Figure \ref{fig:abstract}).
In our study, we (1) compare their diagnostic performance against specialised EEG models across three public datasets, (2) evaluate the necessity of domain adaptation, and (3) analyse their ability to localise demographic and disease-related information.

\section{Materials \& Methods}
\subsection{General-Purpose Models \& Domain Adaptation Strategies}
Our study includes three general-purpose models, namely \texttt{MOMENT} (\textcolor[HTML]{f97306}{$\bullet$}) \cite{Goswami2024}, \texttt{UniTS} (\textcolor[HTML]{f97306}{$\bullet$}) \cite{Gao2024}, and \texttt{OTiS} (\textcolor[HTML]{219ebc}{$\bullet$}) \cite{Turgut2025b}.
These models are pre-trained on large, heterogeneous time series corpora to learn general time series features.
For instance, the most recent \texttt{OTiS} was pre-trained on $640,187$ time series samples from $8$ domains, including $400,000$ ECG ($62.48\,\%$), $203,340$ weather ($31.76\,\%$), $19,614$ audio ($3.06\,\%$), $13,640$ engineering ($2.13\,\%$), $3,367$ EEG ($0.53\,\%$ = $0.42\,\%$ resting-state EEG + $0.11\,\%$ emotion recognition EEG; totalling $125$ recording hours), $115$ economics ($0.02\,\%$), and $111$ banking ($0.02\,\%$) samples.
Moreover, to ensure a fair comparison with specialised EEG models, free from architectural, training, and scaling biases, we additionally pre-train \texttt{OTiS} from sratch exclusively on the $3,367$ EEG samples following \cite{Turgut2025b} and include this specialised variant as \texttt{OTiS}$_\text{EEG}$ (\textcolor[HTML]{b7094c}{$\bullet$}). 

We evaluate three domain adaptation strategies. 
For \textbf{zero-shot} (ZS), the model is frozen after pre-training and evaluated without any fine-tuning. 
Its output tokens are averaged to obtain a global representation. 
Class logits are computed via cosine similarity between a test sample’s representation and each class representation, i.e. the mean global representation of all training samples from a class. 
This adaptation strategy applies only for classification, while the following two also support regression.
For \textbf{linear probing}, the model remains frozen while a randomly initialised linear layer is trained.
For \textbf{fine-tuning} (FT), both the model and a randomly initialised linear layer are trained.

\subsection{Datasets}
Our extensive experiments include three publicly available datasets covering distinct tasks, as detailed in Table \ref{tab:datasets}.
\begin{table}[!t]
\caption{
Overview of datasets used for regression and classification solely from EEG.
}
\label{tab:datasets}
\centering
\footnotesize
\setlength{\tabcolsep}{0.315em}
\begin{tabular}{llrrrrrr}
    \toprule
    \textbf{\normalsize Dataset} & \textbf{\normalsize Task} & \textbf{\normalsize \# Samples} & \textbf{\normalsize \# Variates} & \multicolumn{2}{r}{
    \textbf{\normalsize \# Time points}} & \textbf{\normalsize Frequency} \\ 
    & & & & \textit{per sample} & \textit{total} & \\
    \midrule
    LEMON \cite{Babayan2019} & Age prediction & $378$ & $32$ & $30,000$ & $11\,$M    & $250\,$Hz \\
    Epilepsy \cite{Andrzejak2001} & Seizure detection (Binary) & $11,500$  &  $1$ & $178$ & $2\,$M  & $174\,$Hz \\
    TUEV \cite{Harati2015} & Event classification (Multi-class) & $112,237$ & $19$ & $1,000$ & $112\,$M & $200\,$Hz \\
    \bottomrule
\end{tabular}
\end{table}
\textbf{LEMON} \cite{Babayan2019} comprises resting-state EEG sampled at $250\,$Hz from healthy subjects aged $20-35$ years ($67\,\%$) and $59-77$ years ($33\,\%$).
\textbf{Epilepsy} \cite{Andrzejak2001} includes single-channel EEG from healthy subjects at rest ($20\,\%$) and patients during epileptical seizures ($80\,\%$), sampled at $174\,$Hz and band-pass filtered between $0.5-40\,$Hz.
\textbf{TUEV} \cite{Harati2015} is a large EEG corpus with patient recordings of three clinically relevant events, including 
spike and sharp waves (SPSW: $2\,\%$), 
generalised periodic epileptiform discharges (GPED: $7.06\,\%$), 
and periodic lateralised epileptiform discharges (PLED: $12.58\,\%$),
as well as three noise events, such as
eye movement (EYEM: $1.16\,\%$), 
artifacts from equipment or the environment (ARTF: $7.79\,\%$), 
and background activity (BCKG: $69.41\,\%$).

\subsection{Experimental Setup}
\subsubsection{Processing \& Evaluation.} 
We follow established data processing, splitting, and evaluation protocols for age regression on LEMON \cite{Engemann2022} and classification on Epilepsy \cite{Zhang2022} and TUEV \cite{Yang2024}, reporting results across five random seeds for both linear probing and fine-tuning.
To this end, we measure the coefficient of determination ($R^2$) for regression on LEMON and accuracy (ACC)/balanced accuracy (bACC) for classification on Epilepsy/TUEV. 
Regression and classification tasks are optimised using a mean squared error and cross-entropy loss, respectively. 
Training is performed using early stopping, and the model achieving the best validation performance is evaluated on the test set. 
Optimal hyperparameters are found through a grid search over the learning rate ($3$e-$5$, $1$e-$4$, $3$e-$4$, $1$e-$3$, $3$e-$3$), batch size ($2^x$, $x\in[2, 3, ..., 7]$), drop path ($0.1$, $0.2$), layer decay ($0.5$, $0.75$), weight decay ($0.0$, $0.1$, $0.2$), and label smoothing ($0.0$, $0.1$, $0.2$).
All experiments are conducted on one NVIDIA RTX A$6000$-$48$GB GPU.

\subsubsection{Baselines.} 
We benchmark the general-purpose models against $16$ specialised EEG models ($^\dagger$), including $2$ foundational EEG models ($^\ddagger$), and $4$ statistical feature-based approaches ($^*$). 
For age prediction, we compare against regression toward the mean (RTM; predictions equal the training data's mean age)$^*$, handcrafted features$^*$ \cite{Engemann2022}, the filterbank Riemann model$^*$ \cite{Sabbagh2020}, the filterbank source model$^*$ \cite{Engemann2022}, shallow ConvNet$^\dagger$ \cite{Schirrmeister2017}, and deep ConvNet$^\dagger$ \cite{Schirrmeister2017}.
For seizure detection, we include SimCLR$^\dagger$ \cite{Tang2020}, TimesNet$^\dagger$ \cite{Wu2022}, CoST$^\dagger$ \cite{Woo2022}, TS2Vec$^\dagger$ \cite{Yue2022}, TF-C$^\dagger$ \cite{Zhang2022}, Ti-MAE$^\dagger$ \cite{Li2023}, and SimMTM$^\dagger$ \cite{Dong2024}, all pre-trained on SleepEEG \cite{Kemp2000} totalling $205$ recording hours.
For EEG event type classification, the baselines comprise ST-Transformer$^\dagger$ \cite{Song2021}, CNN-Transformer$^\dagger$ \cite{Peh2022}, FFCL$^\dagger$ \cite{Li2022}, SPaRCNet$^\dagger$ \cite{Jing2023}, ContraWR$^\dagger$ \cite{Yang2023}, BIOT$^\ddagger$ ($3\,$M parameter, pre-trained on $13,000$ recording hours) \cite{Yang2024}, and LaBraM$^\ddagger$ ($370\,$M parameter, pre-trained on $2,500$ recording hours) \cite{Jiang2024}.

\section{Results \& Discussion}
\begin{figure}[!t]
    \centering
    \begin{subfigure}[t]{0.32\textwidth}
        \centering
        \includegraphics[width=0.8\linewidth]{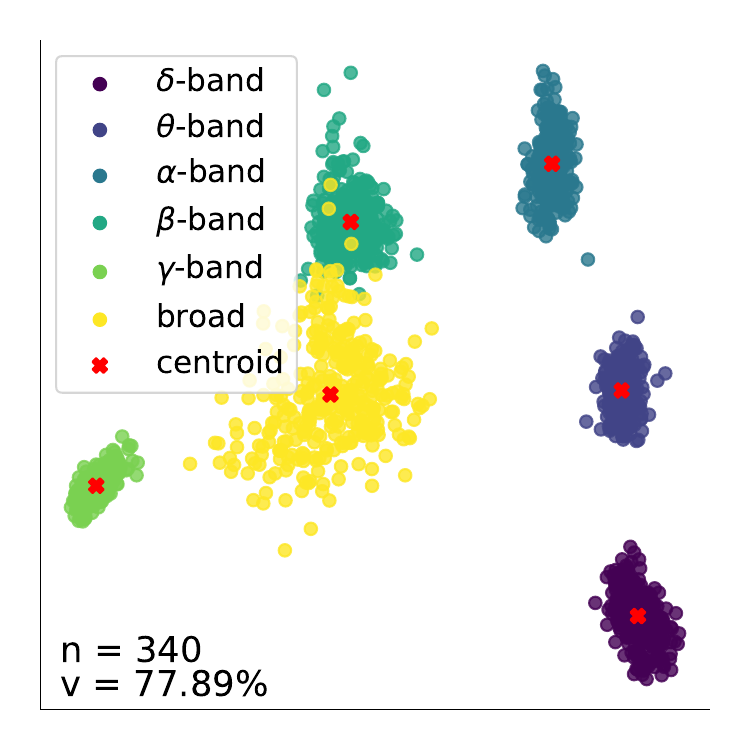} 
        \caption{LEMON.}
        \label{fig:app_zero_frequency}
    \end{subfigure}
    \begin{subfigure}[t]{0.32\textwidth}
        \centering
        \includegraphics[width=0.8\linewidth]{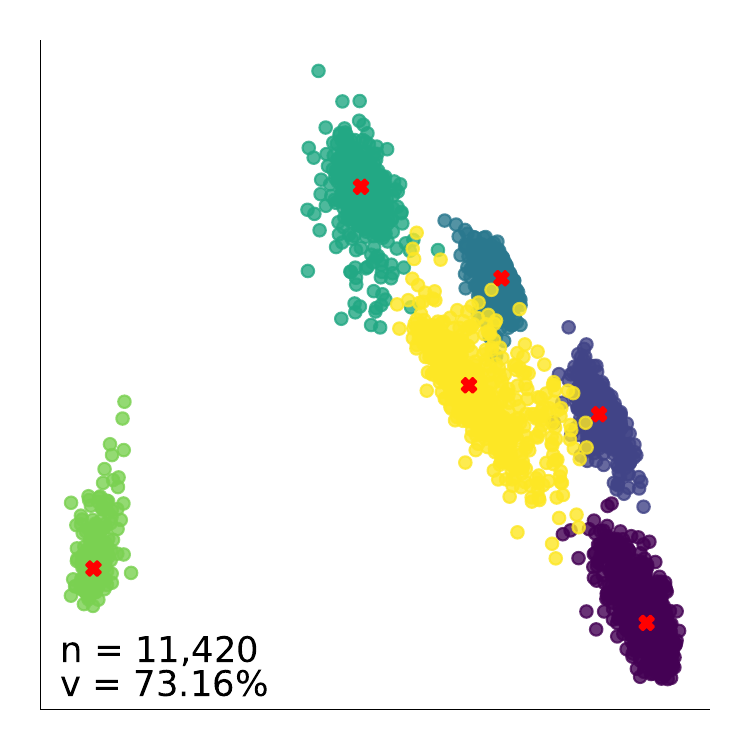}
        \caption{Epilepsy.}
        \label{fig:app_zero_amplitude}
    \end{subfigure}
    \begin{subfigure}[t]{0.32\textwidth}
        \centering
        \includegraphics[width=0.8\linewidth]{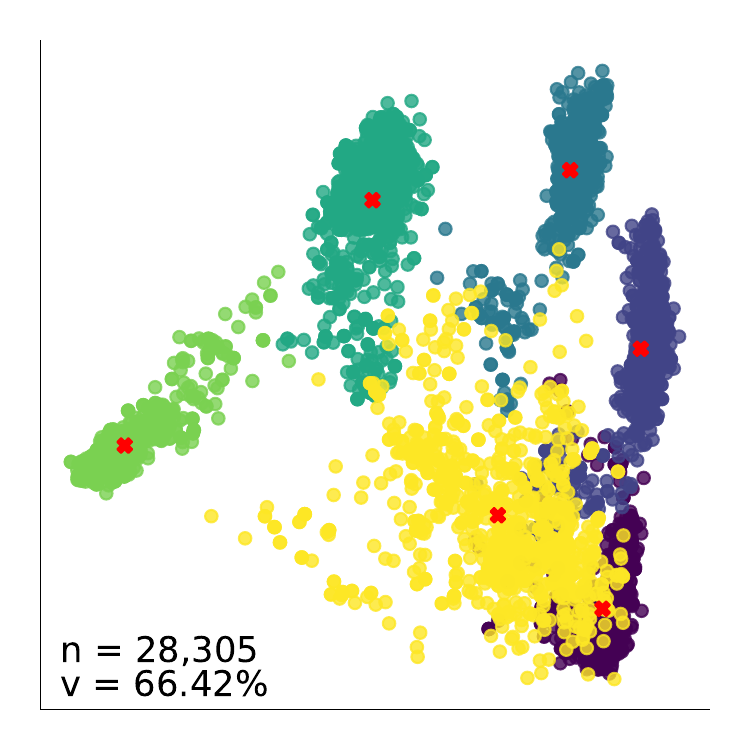}
        \caption{TUEV.}
        \label{fig:app_zero_offset}
    \end{subfigure}
    \caption{
    PCA of zero-shot EEG features.  
    By capturing distinct EEG features across frequency bands, general-purpose models such as \texttt{OTiS} may offer valuable benefits for clinical practice.
    }
    \label{fig:latent_space}
\end{figure}

General-purpose time series models capture distinct EEG features across frequency bands, as visualised in Figure \ref{fig:latent_space}, but their clinical utility remains unclear.
To investigate this, we (1) compare the diagnostic performance of such models against specialised EEG models (Section \ref{sec:performance}), (2) evaluate whether they require domain adaptation to extract clinically relevant information (Section \ref{sec:domain_adaptation}), and (3) analyse their potential to localise demographic and disease-related biomarkers (Section \ref{sec:localisation}).

\subsection{Diagnostic Performance}
\label{sec:performance}
\begin{figure}[!t]
    \centering
    \includegraphics[width=1.0\linewidth]{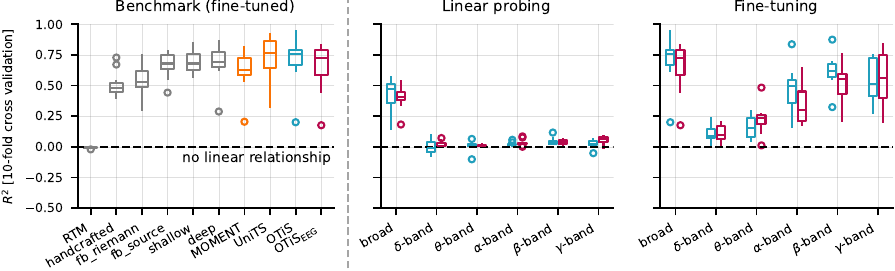}
    \caption{
    \textbf{LEMON}.
    (Left) General-purpose models (\textcolor[HTML]{f97306}{$\bullet$},\textcolor[HTML]{219ebc}{$\bullet$}) are competitive with specialised models (\textcolor{gray}{$\bullet$},\textcolor[HTML]{b7094c}{$\bullet$}).
    (Right) Fine-tuning is essential for clinical utility. 
    Age effects are detected in higher frequencies.
    }
    \label{fig:age_reg}
\end{figure}

\begin{figure}[!t]
    \centering
    \includegraphics[width=1.0\linewidth]{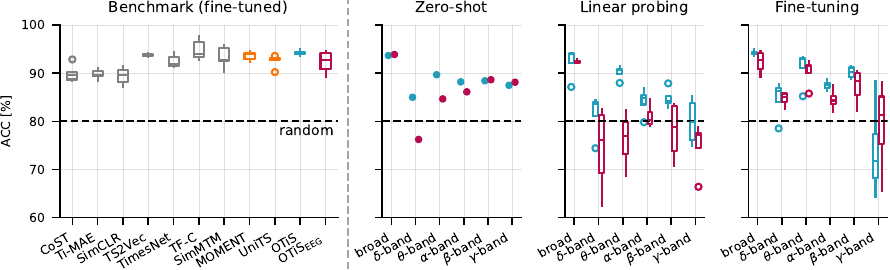}
    \caption{
    \textbf{Epilepsy}.
    (Left) General-purpose models (\textcolor[HTML]{f97306}{$\bullet$},\textcolor[HTML]{219ebc}{$\bullet$}) are competitive with specialised models (\textcolor{gray}{$\bullet$},\textcolor[HTML]{b7094c}{$\bullet$}).
    (Right) Domain adaptation through linear probing or fine-tuning is not required for seizure detection. Seizure-related information shows no clear frequency localisation.
    }
    \label{fig:epilepsy_cls}
\end{figure}

We study whether general time series knowledge, obtained through pre-training on large and heterogeneous time series corpora, offers advantages for EEG analysis.
To this end, we compare three general-purpose models - \texttt{MOMENT} ($40\,$M) \cite{Goswami2024}, \texttt{UniTS} ($8\,$M) \cite{Gao2024}, and \texttt{OTiS} ($7\,$M) \cite{Turgut2025b} - against specialised EEG models 
trained solely on domain-specific data.
Across benchmarks in age prediction, seizure detection, and EEG event classification, general-purpose models perform competitively, surpassing statistical baselines and in some cases even specialised models (see Figures \ref{fig:age_reg}, \ref{fig:epilepsy_cls}, and \ref{fig:tuev_cls}, left).
Figure \ref{fig:tuev_cls} suggests that general-purpose models might even capture clinically relevant information beyond what is achieved by specialised foundation models such as BIOT ($3\,$M) \cite{Yang2024}.
Comparisons with the specialised \texttt{OTiS}$_\text{EEG}$ ($7\,$M) further highlight the contribution of general time series pre-training, while ruling out differences in architecture and model size.
Large-scale domain-specific pre-training can yield optimal performance, as indicated by LaBraM \cite{Jiang2024} in Figure \ref{fig:tuev_cls}, but such approaches are often limited by data availability. 
Overall, our findings suggest that general-purpose models can reduce reliance on large domain-specific datasets while retaining competitive diagnostic performance.

\subsection{Domain Adaptation Strategies}
\label{sec:domain_adaptation}
\begin{figure}[!t]
    \centering
    \begin{subfigure}[t]{0.243\textwidth}
        \centering
        \includegraphics[width=0.9\linewidth]{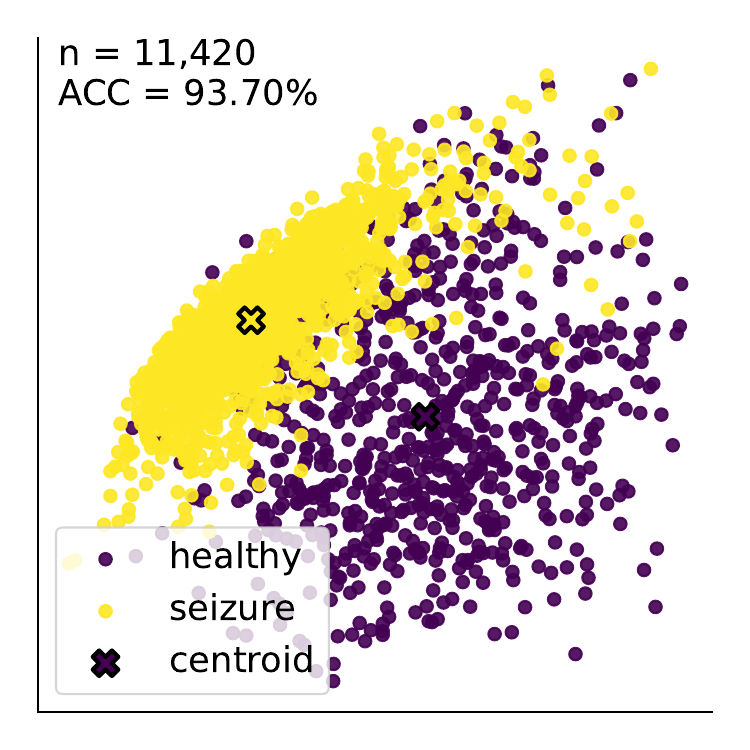}
        \caption{Epilepsy ZS.}
        \label{fig:epilepsy_zero}
    \end{subfigure}
    \begin{subfigure}[t]{0.244\textwidth}
        \centering
        \includegraphics[width=0.9\linewidth]{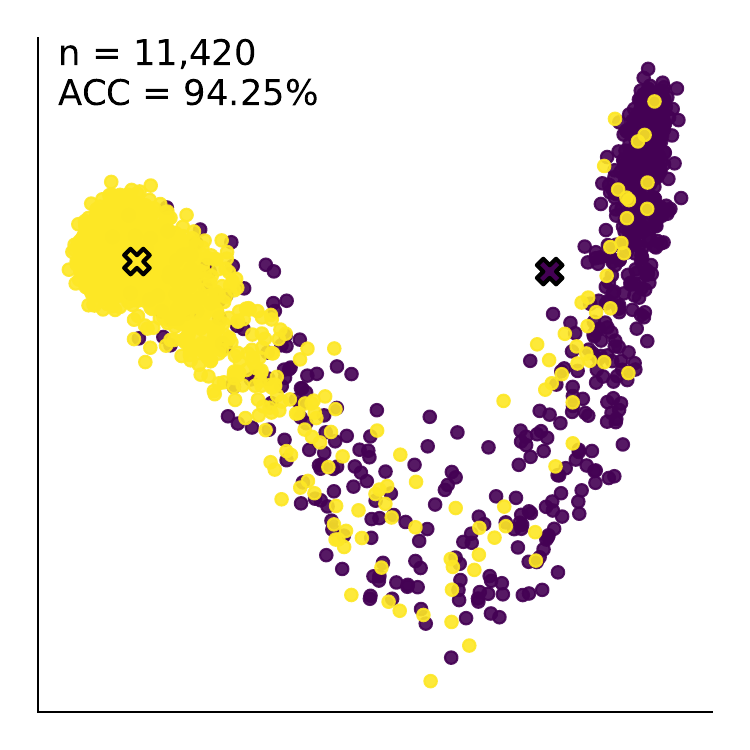}
        \caption{Epilepsy FT.}
        \label{fig:epilepsy_ft}
    \end{subfigure}
    \begin{subfigure}[t]{0.243\textwidth}
        \centering
        \includegraphics[width=0.9\linewidth]{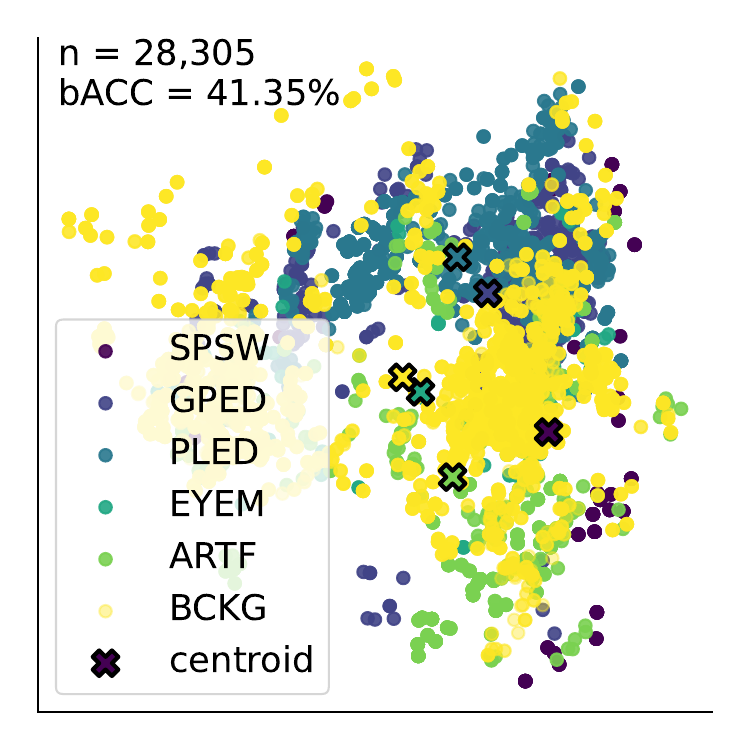}
        \caption{TUEV ZS.}
        \label{fig:tuev_zero}
    \end{subfigure}
    \begin{subfigure}[t]{0.244\textwidth}
        \centering
        \includegraphics[width=0.9\linewidth]{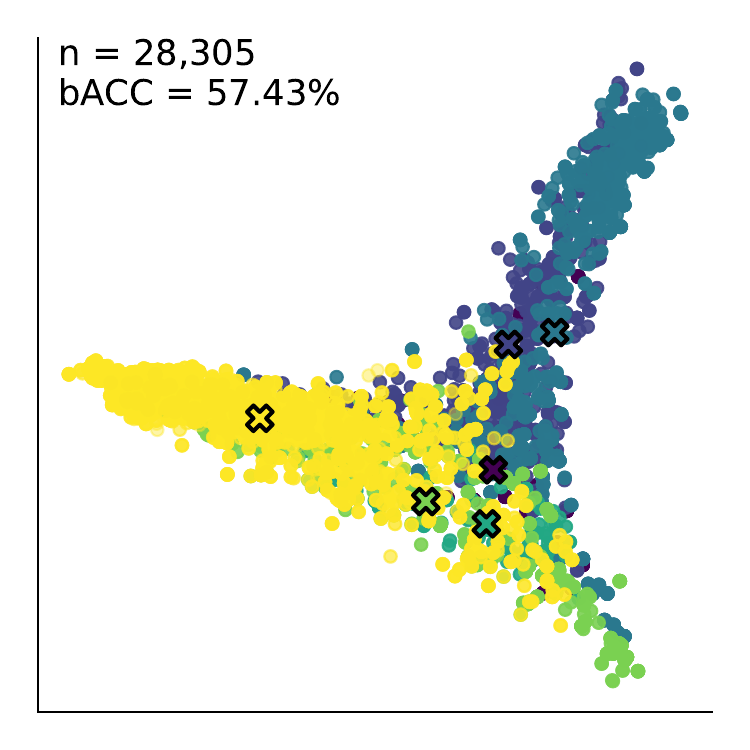}
        \caption{TUEV FT.}
        \label{fig:tuev_ft}
    \end{subfigure}
    \caption{
    PCA of EEG features extracted by \texttt{OTiS}.
    (a, b) Distinct features are captured for healthy subjects and patients, even in the zero-shot (ZS) setting.
    (c, d) The extraction of clinically relevant features is substantially enhanced with fine-tuning (FT).
    }
    \label{fig:cls_embeddings}
\end{figure}

We analyse whether general-purpose models can be applied to EEG analysis out-of-the-box or require domain adaptation.
Therefore, we evaluate \texttt{OTiS} (\textcolor[HTML]{219ebc}{$\bullet$}) and the specialised \texttt{OTiS}$_\text{EEG}$ (\textcolor[HTML]{b7094c}{$\bullet$}) under zero-shot, linear probing, and fine-tuning settings (see \textit{broad} in Figures \ref{fig:age_reg}, \ref{fig:epilepsy_cls}, and \ref{fig:tuev_cls}, right).
Our experiments span datasets of varying scale: $11\,$M time points in LEMON, $2\,$M in Epilepsy, and $112\,$M in TUEV. 
We find that large domain-specific datasets, such as LEMON and TUEV, enable substantial improvements in feature quality through fine-tuning, whereas task-specific training on limited data, as in Epilepsy, risks performance loss through overfitting.
Age prediction in LEMON proves particularly challenging, as evidenced by the linear probing results, and requires task-specific fine-tuning for clinical usability.
For tasks where a visual interpretation of raw EEG is feasible, features of general-purpose models can be used out-of-the-box, as indicated by competitive zero-shot performances in Epilepsy and TUEV.
In particular, such readily available features are useful to detect spike and sharp waves (SPSW) indicative of epileptic seizures (see Figure \ref{fig:epilepsy_zero}), or to distinguish SPSW from eye movement (EYEM) (see Figure \ref{fig:tuev_zero}). 
Fine-tuning offers no advantages for simple tasks such as seizure detection (see Figures \ref{fig:epilepsy_zero} and \ref{fig:epilepsy_ft}) but is essential for complex tasks that require domain knowledge such as distinguishing EYEM from background activity (BCKG) (see Figures \ref{fig:tuev_zero} and \ref{fig:tuev_ft}).
Finally, pre-training on large-scale heterogeneous time series data proves beneficial if domain-specific pre-training data is limited, as demonstrated by the comparison between \texttt{OTiS} and its specialised counterpart \texttt{OTiS}$_\text{EEG}$. 

\subsection{Biomarker Localisation} 
\label{sec:localisation}
\begin{figure}[!t]
    \centering
    \includegraphics[width=1.0\linewidth]{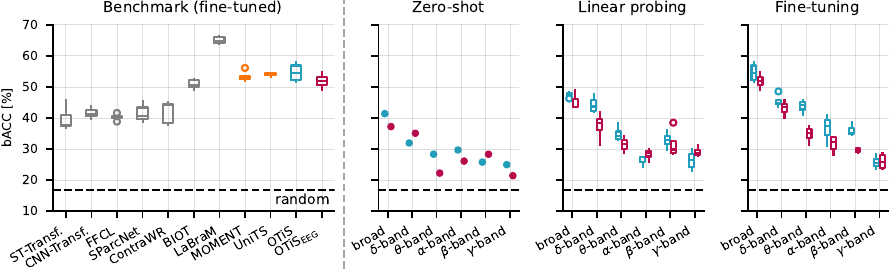}
    \caption{
    \textbf{TUEV}.
    (Left) General-purpose models (\textcolor[HTML]{f97306}{$\bullet$},\textcolor[HTML]{219ebc}{$\bullet$}) outperform nearly all specialised models (\textcolor{gray}{$\bullet$},\textcolor[HTML]{b7094c}{$\bullet$}).
    (Right) Zero-shot features are already expressive, but fine-tuning improves feature quality.
    Clinically relevant markers of seizure or acute neurological conditions are concentrated in lower frequencies.
    }
    \label{fig:tuev_cls}
\end{figure}

To evaluate whether general-purpose models enable the localisation of clinical biomarkers, we assess the expressiveness of EEG features across frequency bands (see Figures \ref{fig:age_reg}, \ref{fig:epilepsy_cls}, and \ref{fig:tuev_cls}, right).
Optimal predictions are consistently obtained using broadband features, suggesting that they capture the full spectrum of clinically relevant information.
Assessing predictions within individual frequency bands allows localisation of specific biomarkers: age-related information is encoded in higher frequencies (Figure \ref{fig:age_reg}), whereas ictal activity is concentrated in lower frequencies (Figure \ref{fig:tuev_cls}).
These findings align with literature on age-related EEG biomarkers \cite{Hashemi2016, Kang2024} and epileptical markers \cite{Panet1990, Gaspard2013}.
Band-specific analyses further reveal that data pre-processing heavily affects diagnostic performance: radical low-pass filtering of raw EEG at $40\,$Hz \cite{Andrzejak2001} reduces $\gamma$-band seizure detection to chance levels in Epilepsy ($80\%$ majority class; Figure \ref{fig:epilepsy_cls}).
Interestingly, zero-shot features (Figure \ref{fig:latent_space}) already reflect these findings: broadband features align with features of the most informative bands (LEMON: $\beta$-band; TUEV: $\delta$-band) and diverge from the ones of less informative bands (Epilepsy: $\gamma$-band).


\section{Conclusion}
In this study, we explore the utility of general-purpose time series models for electroencephalography (EEG) analysis.
Through extensive benchmarking on age prediction, seizure detection, and EEG event type classification, we find that these models are competitive with specialised EEG models. 
While general-purpose models can be used out-of-the-box for tasks where a visual interpretation of raw EEG is feasible (e.g. seizure detection), domain adaptation through fine-tuning becomes essential for tasks that require higher level of specialisation (e.g. event type classification).
Furthermore, our experiments reveal that these models enable the localisation of demographic and disease-specific information through frequency band analysis.
These findings indicate that foundation models can be useful in clinical routine, especially when domain-specific data is limited.
Overall, we believe this work provides valuable guidance for integrating general-purpose models into clinical practice and motivates future exploration in other EEG tasks, such as sleep stage classification or motor imagery.

\bibliography{neurips_2025.bib}
\bibliographystyle{abbrvnat}

\end{document}